\documentclass[10pt]{article}

\usepackage{arxiv}

\usepackage[utf8]{inputenc} 
\usepackage[T1]{fontenc}    
\usepackage{hyperref}       
\usepackage{url}            
\usepackage{booktabs}       
\usepackage{amsfonts}       
\usepackage{nicefrac}       
\usepackage{microtype}      
\usepackage{lipsum}
\usepackage{graphicx}
\usepackage{subcaption}

\usepackage{mathptmx}
\usepackage{times}

\graphicspath{ {./images/} }

\title{Anomaly Detection with SDAE}

\author{
 Benjamin Smith \\
  University of Victoria\\
  \texttt{benjaminsmith@uvic.ca} \\
   \And
 Kevin Cant \\
  University of Victoria\\
  \texttt{kcant@uvic.ca} \\
  \And
 Gloria Wang \\
  University of Victoria\\
  \texttt{gwang@uvic.ca} \\
}

\begin{document}
\maketitle
\begin{abstract}
Anomaly detection is a prominent data preprocessing step in learning applications for correction and/or removal of faulty data. 
Automating this data type with the use of autoencoders could increase the quality of the dataset by isolating anomalies that were missed through manual or basic statistical analysis. 
A Simple, Deep, and Supervised Deep Autoencoder were trained and compared for anomaly detection over the ASHRAE building energy dataset. 
Given the restricted parameters under which the models were trained, the Deep Autoencoder perfoms the best, however, the Supervised Deep Autoencoder outperforms the other models in total anomalies detected when considerations for the test datasets are given.
\end{abstract}


\section{Introduction}

Autoencoders (AE), prominent in the space of dimensional reduction, are typically trained in an unsupervised manner. 
They are used to map inputs to a representational encoding that describes the latent space of the input distribution. 
Accuracy improvements have been made by making these AE models deeper, leading to Deep AE (DAE). 
Introducing a supervised loss, typically optimized by stochastic gradient descent (SGD), in conjunction with the reconstruction loss provides better extraction representations that are tailored to class labels, and thus we have Supervised Deep AutoEncoders (SDEA). 

Autoencoders are capable of taking a feature space and reducing its dimensionality into a representative latent space that captures the important criteria of the input. 
With the ability to create a reduced dimensionality space, autoencoders have been used successfully in the realm of anomaly detection \cite{AnomalyDetectionUsingAutoencodersWithNonlinearDimensionalityReduction}.
Anomaly detection is the method of discovering outliers within a distribution. 
These outliers can occur from many types of situations such as sensor corruption or fault, incorrect manually entered data, or faulty processes, for example. 
We are particularly interested in the sensor corruption/fault anomalies as these are labour intensive to discover manually when large amounts of data are being gathered at once or as a stream. 

\section{Anomaly Detection} 

Data pre-processing for real-world applications forms a significant amount of time in the overall software life-cycle \cite{Visual-InteractivePreprocessing}.
This process involves removing or correcting anomalous entries and though some may be obvious, there are possible anomalies that can go undetected by conventional manual or statistical analysis. 
These anomalies, say in terms of machine learning, if numerous enough can skew a model's the ability to learn the underlying distribution. 
This skew leads to improper application usage that could potentially be harmful depending on the use case (genome feature detection in cancer patients for example). 

Anomaly detection also presents itself as a unique problem because of the imbalance in class labels between expected and anomalous examples. This imbalance requires the use of well-suited loss functions \cite{variationalAE}.

\section{Related Works}

Before autoencoders were introduced in to the anomaly detection domain, PCA systems were the best approach \cite{AnomalyDetectionUsingAutoencodersWithNonlinearDimensionalityReduction}.
However, autoencoders were shown to outperform PCAs and thus the focus shall be on autoencoders \cite{AnomalyDetectionUsingAutoencodersWithNonlinearDimensionalityReduction}.
Autoencoder training is typically done via the reconstruction loss.
Once the autoencoder has taken in an input example, compressed it into its latent space, and then reconstructed it into the final prediction, the reconstruction loss determines how dissimilar the prediction is from the original input. 
The smaller the loss, the more accurate the reconstruction is \cite{AnomalyDetectionUsingAutoencodersWithNonlinearDimensionalityReduction}. 
A variety of autoencoder types have been proposed in literature, including a normal autoencoder, a denoising \cite{AnomalyDetectionUsingAutoencodersWithNonlinearDimensionalityReduction}, deep autoencoder \cite{Usingdeepautoencoderstoidentifyabnormalbrainstructuralpatternsinneuropsychiatricdisorders}, semi-supervised autoencoder \cite{Usingdeepautoencoderstoidentifyabnormalbrainstructuralpatternsinneuropsychiatricdisorders}, and variational autoencoders \cite{FaceValidationBasedAnomalyDetectionUsingVariationalAutoencoder}.

Applications of anomaly detection that have AE applied to them with success involve face validation \cite{FaceValidationBasedAnomalyDetectionUsingVariationalAutoencoder}, satellite data validation \cite{AnomalyDetectionUsingAutoencodersWithNonlinearDimensionalityReduction}, identifying abnormal brain structural patterns in neuropsychiatric disorders \cite{Usingdeepautoencoderstoidentifyabnormalbrainstructuralpatternsinneuropsychiatricdisorders}, vocal deceptions \cite{Improvedsemi-supervisedautoencoderfordeceptiondetection}, industrial optical inspection \cite{Semi-supervisedAnomalyDetectionUsingAutoEncoders}, and rail surface discrete defects \cite{Semi-supervisedAnomalyDetectionUsingAutoEncoders} as examples. 
It is important to note that the data included in the aforementioned examples consisted of an imbalanced class ratio of anomalies to normal data, yet AE's are still able to pick up the anomalous entries. 
The practice approach is to train an AE on the normal data and verify anomalies via the reconstruction loss value exceeding a particular threshold latent to its distribution \cite{Semi-supervisedAnomalyDetectionUsingAutoEncoders, AnomalyDetectionUsingAutoencodersWithNonlinearDimensionalityReduction, FaceValidationBasedAnomalyDetectionUsingVariationalAutoencoder, Improvedsemi-supervisedautoencoderfordeceptiondetection}.
The anomaly score, based on the reconstruction error, is shown in Eq \ref{anomaly_score}

\begin{equation}
    Err_{i} = \sqrt{\Sigma_{j=1}^{D} (x_{j}(i) - \hat{x}_{j}(i))^{2} }
    \label{anomaly_score}
\end{equation}
where $x_{i}$ is the input and $\hat{x}_{i}$ is the output of the model and $i \epsilon [1,...,|D|]$ \cite{AnomalyDetectionUsingAutoencodersWithNonlinearDimensionalityReduction}. 

Experiments have shown that including a supervised component to an AE has increased the identification of anomalies \cite{Improvedsemi-supervisedautoencoderfordeceptiondetection}. 
Having labeled anomalous data enables a distinct decision boundary to exist in the distribution - making classifications of anomalous entries more precise. 
This mapping between input and output classes can extract better representations that are tailored to said classes \cite{JointlyPre-trainingwithSupervisedAutoencoderandValueLossesforDeepReinforcementLearning}. 
Integrating a supervised loss in with the reconstruction loss aids in such a refined mapping as shown from \cite{JointlyPre-trainingwithSupervisedAutoencoderandValueLossesforDeepReinforcementLearning} in Eq \ref{supervised_loss}.
\begin{equation}
    L^{sae} = L_{s}^{sae}(W_{s} F(x_{i}, y_{i})) + L_{ae}^{sae}(W_{ae} F(x_{i}, \hat{x}_{i}))
    \label{supervised_loss}
\end{equation}

\section{Methodology}

\subsection{Dataset}

The dataset from the Kaggle competition ASHRAE is used \cite{kaggle}. 
The dataset provides the features seen in Table \ref{tab:dataset_features} that are used in training.

\begin{table}[h]
    \centering
    \begin{tabular}{|c|c|}
    \hline
        Feature & Description \\
    \hline
        building\_id & Foreign key for the building metadata \\
    \hline
        square\_feet & Floor area of the building\\
    \hline
        year\_built & Build year of the building\\
    \hline
        floor\_count & Number of floors of the building\\
    \hline
        air\_temperature & Outside air temperature in Celsius \\
    \hline
        cloud\_coverage & Portion of sky covered\\
    \hline
        dew\_temperature & Dew temperature in Celsius \\
    \hline
        precip\_depth\_1\_hr & How much precipitation occurred in the last hour\\
    \hline
        sea\_level\_pressure & Pressure above sea level in millibar units \\
    \hline
        wind\_direction & Direction of wind in 360$^{o}$, where 0$^{o}$ is North\\
    \hline
        wind\_speed & Speed of wind in meters per second \\
    \hline
        meter & The meter id code. Read as {0: electricity, 1: chilled water, 2: steam, 3: hot water}\\
    \hline
        meter\_reading & Value read on the meter at a given time in kWh \\
    \hline
    \end{tabular}
    \caption{Dataset Features}
    \label{tab:dataset_features}
\end{table}{}

The source datasets of "train.csv", "building\_metadata.csv" 
and "weather\_train.csv" were combined into 
"Cleaned\_Training\_Data.csv" which has the shape (20216100, 17). 
Since the focus on supervised learning, this dataset was then subset into a labeled training set that contained all the information from Buildings 18, 22, 40, 49, and 490, which were selected at random. 
This new training dataset was of shape (40000, 14) where the first 13 features were trainable data and the last was the anomaly label: 1 for anomalous and 0 for non-anomalous. Anomaly labelling was completed manually via domain expertise. 
The dataset contained over 1,000 individual buildings, with up to four meters per building, recorded hourly for a year (approximately 8,000 datapoints per meter per building). 
The time required for manual removal of anomalous data points on an hour-by-hour basis is infeasible. 

Seen in Table \ref{tab:dataset} are the breakdown of features within the source dataset.
Buildings were built within the range of 1960 to 2020 and favoured smaller square footage.
Since energy consumption depends on the age of the building (dependent on renovations - but this metric was not tracked), square footage, and outside temperature - these features were key to training. 
It can be further suggested that air temperature is dependent on the sea level pressure, wind direction and speed, cloud coverage, and precipitation.
These dependencies were the reason the 13 features above were chosen. 

\begin{table}[h]
    \centering
    \begin{tabular}{cccc}
    \subcaptionbox{Number of data in each year\label{1}}{\includegraphics[width = 1.25in]{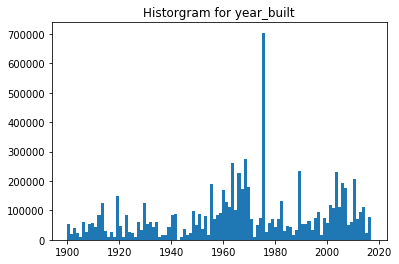}} & 
    
    \subcaptionbox{The square feet for dataset\label{2}}{\includegraphics[width = 1.25in]{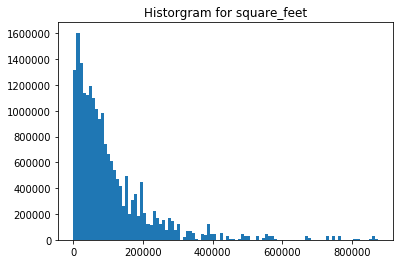}} &
    
    \subcaptionbox{The box plot for square feet\label{3}}{\includegraphics[width = 1.25in]{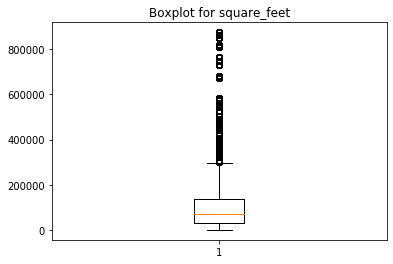}} &
    
    \subcaptionbox{Histogram of temperature\label{4}}{\includegraphics[width = 1.25in]{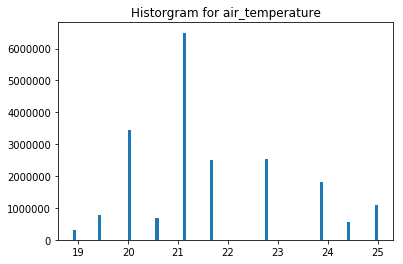}} \\
    
    \subcaptionbox{Box plot for temperature\label{4}}{\includegraphics[width = 1.25in]{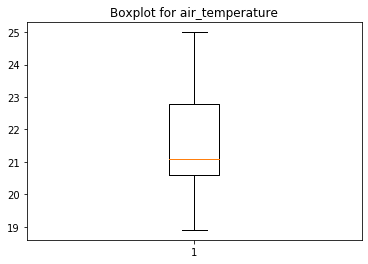}} &
    
    \subcaptionbox{Primary uses for different buildings\label{4}}{\includegraphics[width = 1.25in]{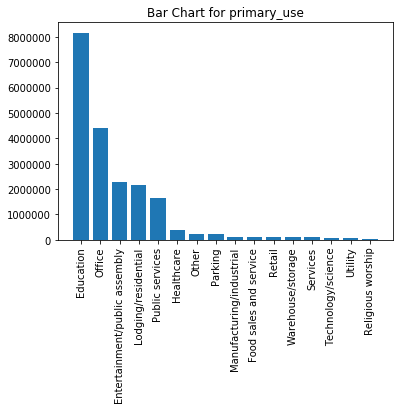}} &
    
    \subcaptionbox{Density during weekdays\label{4}}{\includegraphics[width = 1.25in]{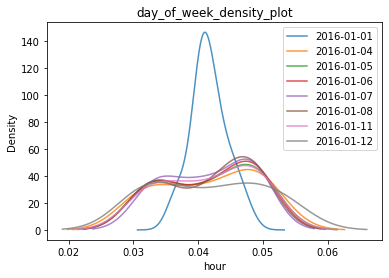}} &
    
    \subcaptionbox{Density during weekends\label{4}}{\includegraphics[width = 1.25in]{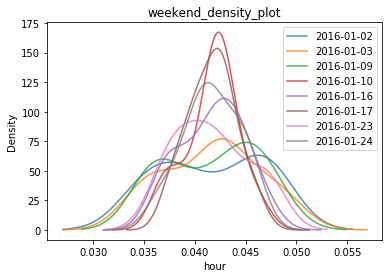}} \\
    
    \end{tabular}
    \caption{Dataset}
    \label{tab:dataset}
\end{table}{}

The distribution of building types, sizes, ages, and locations results in an extensive range of values and simple trends are not readily identifiable. 
The ability for a model to detect anomalies is not trivial and simple algorithms or statistical measures are not suitable for this problem. 

\subsection{Models}

Autoencoders' ability to learn the underlying distribution make them a well-suited choice for anomaly detection. 
The architecture of autoencoders are compromised of two segments: an encoder and a decoder. 
The encoding segment compresses the input, $\mathcal{X}$, feature dimension into a representational latent space that is smaller than the input dimension.
This is achieved through a number of hidden layers within the model. 
The number of hidden layers dictate the depth and complexity of the input distribution that can be captured by the encoding process. 
The decoder reflects the encoder, typically in a symmetrical fashion with the same number of hidden layers. 
The decoder takes the compressed latent space representation and decompresses it back into the dimension of the input, $\hat{\mathcal{X}}$. 
This process of encoding and decoding is used to remove noise, identify important features, or in this study, detect anomalies. 

\subsubsection{Simple Autoencoder}

\begin{figure}[h]
    \centering
    \includegraphics[scale=0.25]{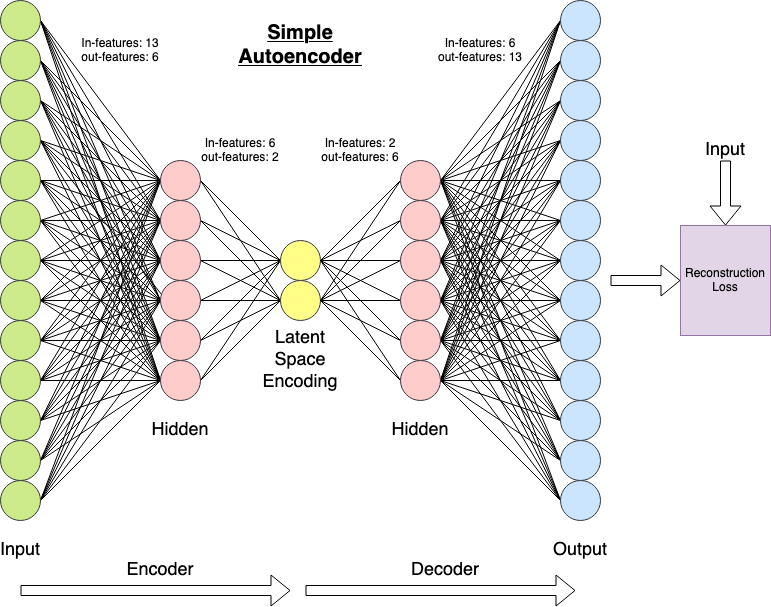}
    \caption{Simple AE}
    \label{fig:simple_ae}
\end{figure}

The Simple AE in Figure \ref{fig:simple_ae} reduces the input to a hidden layer of six features.
This hidden layer is then reduced to the compressed latent space of two features. 
The decoder rebuilds from this space and outputs with 13 features. 
This AE output is assessed with the source input via the reconstruction loss. 

{}

\newpage

\subsubsection{Deep Autoencoder}

\begin{figure}[h]
    \centering
    \includegraphics[scale=0.175]{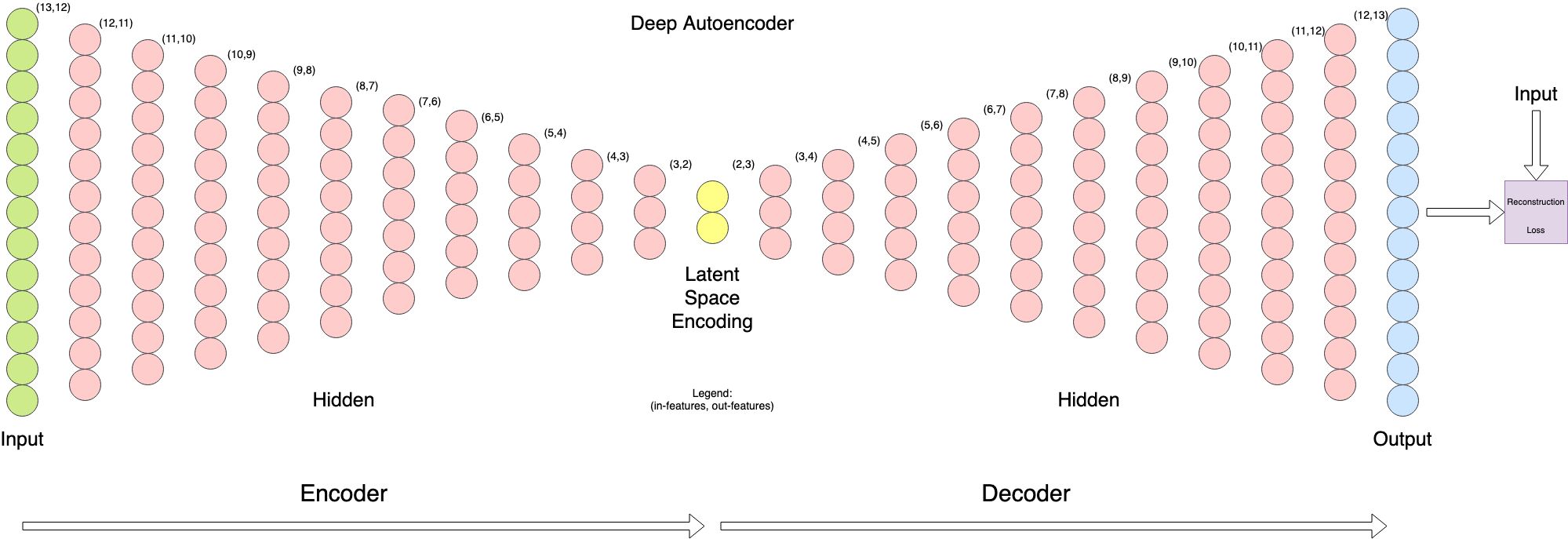}
    \caption{Deep AE}
    \label{fig:deep_ae}
\end{figure}

The Deep AE is just like the Simple AE except it contains many more hidden layers as seen in Figure \ref{fig:deep_ae}.
The Deep AE takes the input and progressively compresses it by one less dimension in the feature space as the previous layer down until the latent space of two features. 
The decoder rebuilds this distribution encoding into an output of the same dimension as the input. 
This AE output is assessed with the source input via the reconstruction loss. 

{}

\subsubsection{Supervised Deep Autoencoder (SDAE)}

The SDAE is almost identical to the Deep AE except for what happens at the compressed latent space layer.
This latent space encoding is passed to a supervised loss. 
The decoder rebuilds from the latent space and passes the output to the reconstruction loss along with the source input. 
The reconstruction loss is paired with the supervised loss as a combined loss that is then back propagated through the model during training. 

\begin{figure}[h]
    \centering
    \includegraphics[scale=0.175]{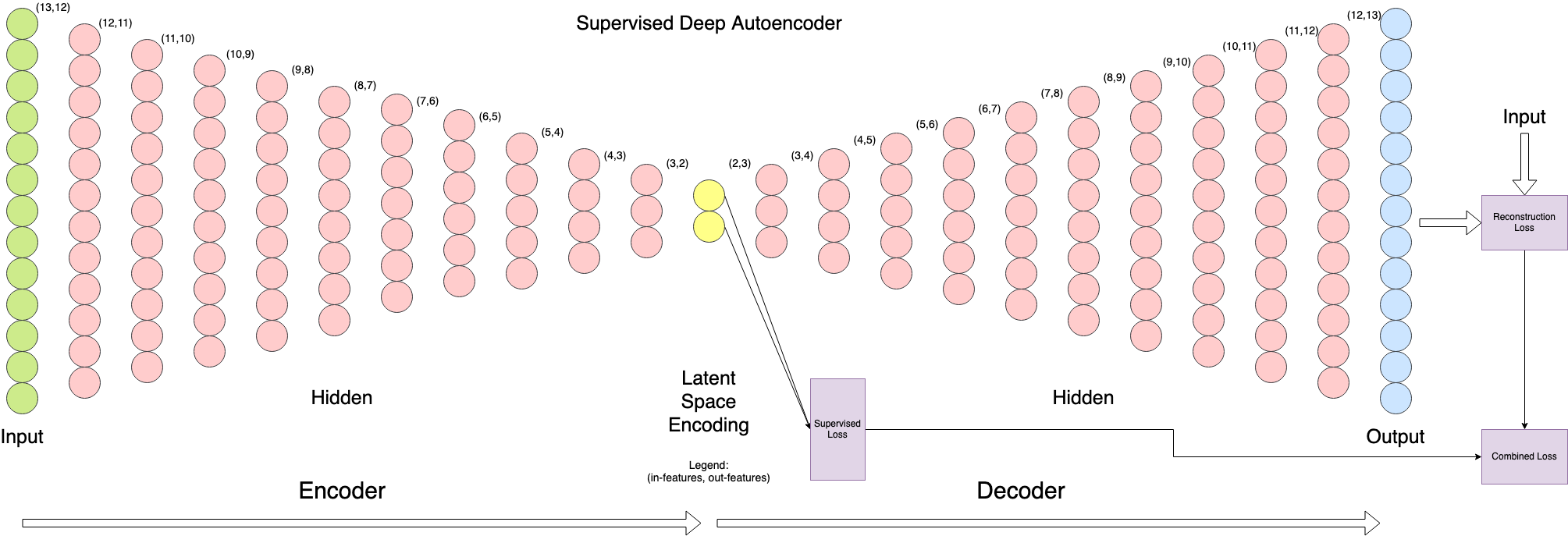}
    \caption{SDAE}
    \label{fig:sdae}
\end{figure}{}

\subsection{Losses}

Typically AEs are unsupervised models, meaning that they do not usually make use of any labeling against outputs. 
However, in this case of the SDAE, a supervised loss is being implemented in order to refine the loss propagation throughout the model with respect to class-imbalanced anomalies. 
While the Simple and Deep AE made use of the reconstruction loss alone, the SDAE uses the supervised loss in combination with the reconstruction loss to further refine the back propagation during training. 

\subsubsection{Reconstruction Loss}

The reconstruction loss is used to measure the average squared distance between two datapoints: the input and the reconstructed input from the AE decoder.
In this case the Mean Square Error (MSE) is used. 

\begin{equation}
    MSE = \frac{1}{n}\Sigma_{i=1}^{n}(\mathcal{X} - \hat{\mathcal{X}}^{2}
    \label{eqn:mse}
\end{equation}{}

\subsubsection{Supervised Loss}

A supervised loss is one that calculates the error of the prediction class possibility to the labeled class it is compared to. 
In this case, the use of Binary Cross Entropy (BCE) was used since we only required labels for two classes: anomalies and non-anomalies. 

\begin{equation}
    BCE = -(y log(p) + (1-y)log(1-p))
    \label{eqn:bce}
\end{equation}{} 

This supervised loss in the SDAE took the shape of a tuple: (1,0) for anomalous data and (0,1) for non-anomalous data. 
This labeling syntax was chosen to emphasize weighting according to the two-feature latent space the SDAE model encoded the input to. 

\section{Results}

The training of the three AE models was performed on a Standard NC24 Microsoft Azure Virtual Machine (VM). 
This VM contained 24 Haswell CPU's and 4 NVIDIA Tesla K80 GPU's. 
The autoencoder training was implemented using  PyTorch's DataParallel wrapper. 
This wrapper implores a scatter-gather technique at distributing the input batch. 
First each model is copied onto each individual GPU. 
The batch is then evenly divided into the number of GPU's - a batch size of 32 was chosen and the resulting batch size at each GPU was 8. 
The reason the batch size was kept small was that larger batch sizes observed poor quality learning with the respect to the loss function - it was erratic and of higher values than with a small batch size.
A more erratic loss value per epoch suggests correlations being made that do not represent effectively the underlying distribution. 
The validation loss was tracked and offered no meaningful information - it stayed constant - so it was dropped from the plots so the training loss could be discussed in observable detail. 

Each model was trained for 1000 epochs and their training times can be seen in Table \ref{training_time_1000ep}. 
As the complexity of the model increases, so does its training time - this is expected and obvious. 
Training cycles are set at 1000 epochs as a reasonable time limits in model development. 

The training data involved a subset of the original dataset focused on Building 18, 22, 40, 49, and 490 totalling 40,000 labeled datapoints. 

\begin{table}[h]
    \begin{center}
        \begin{tabular}{ |c|c| } 
            \hline
            Model & Training Time \\
            \hline
            Simple AE & 3:58:07.777532 \\ 
            \hline
            Deep AE &  8:21:22.935151 \\ 
            \hline
            SDAE & 8:53:44.365703 \\ 
            \hline
        \end{tabular}
    \end{center}
    \captionof{table}{Training over 1000 epochs with batch size 32}
    \label{training_time_1000ep}
\end{table}

\begin{figure}
\centering
\begin{minipage}{.5\textwidth}
  \centering
  \includegraphics[width=\linewidth]{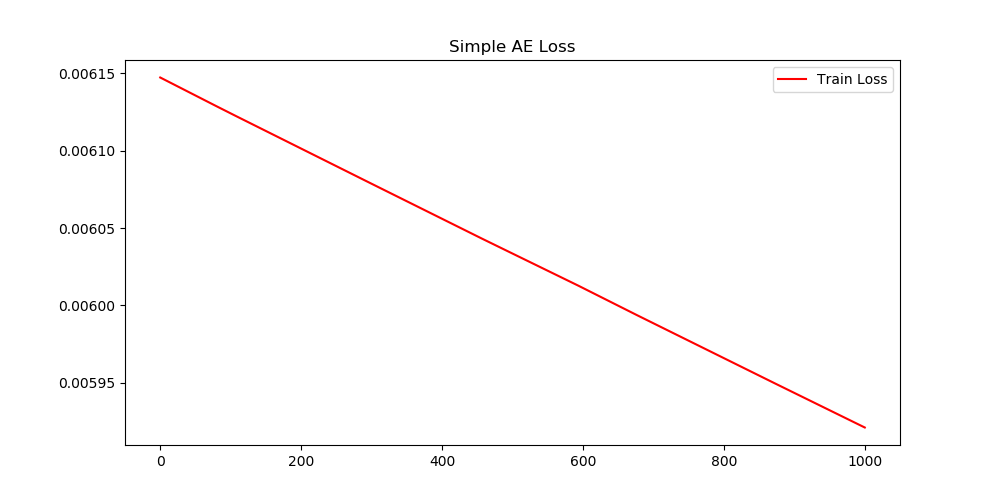}
  \captionof{figure}{Simple AE Loss}
  \label{fig:simple_ae_loss}
\end{minipage}%
\begin{minipage}{.5\textwidth}
  \centering
  \includegraphics[width=\linewidth]{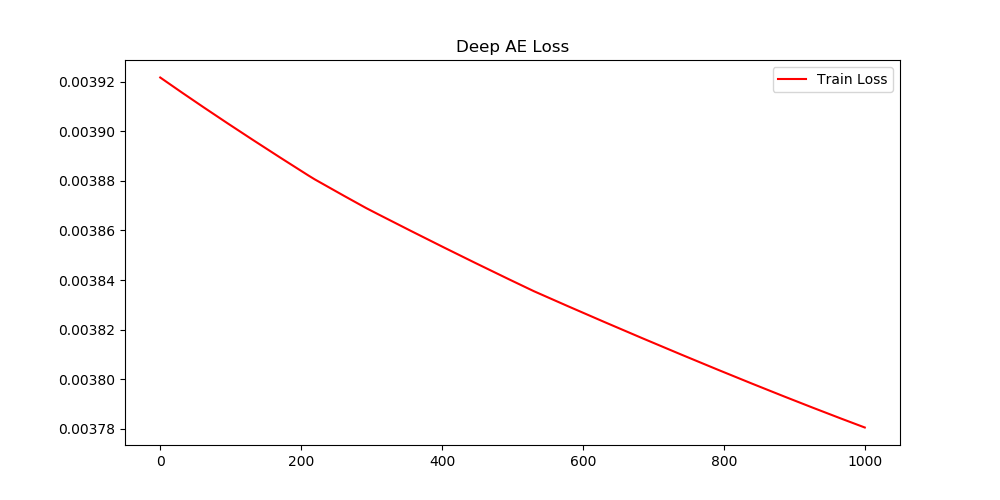}
  \captionof{figure}{Deep AE Loss}
  \label{fig:dae_loss}
\end{minipage}
\end{figure}

\subsection{Simple AE}

The reconstruction loss for the Simple AE can be seen in Figure \ref{fig:simple_ae_loss}. 
The loss values are observed to decrease which implies that the model is learning the underlying distribution and managing to successfully build a reduced dimensional latent space. 
However, the loss function does not level out which implies the model did not converge after 1000 epochs - it is undetermined how many epochs were necessary for a model to converge to a solution. 

\subsection{Deep AE}

The reconstruction loss for the Deep AE can be seen in Figure \ref{fig:dae_loss}.
Like the Simple AE, the loss value decreases which implies a successful learning of the input distribution.
However, there is a more prominent curve which suggests more intricate complexities being captured earlier on in the learning process. 
Again, like the Simple AE, this model did not converge to a solution and it is unknown how long it would need to do so. 

\subsection{SDAE}

\begin{figure}[h]
    \centering
    \includegraphics[scale=0.45]{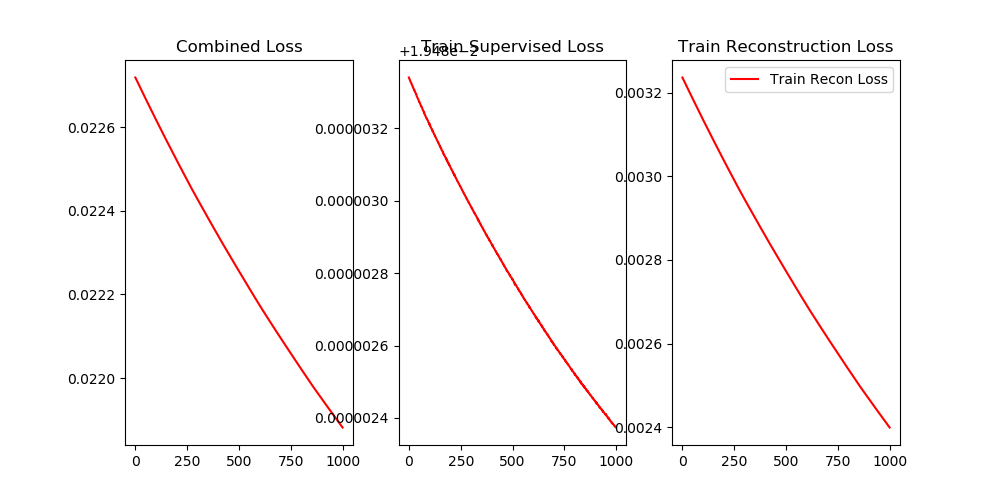}
    \caption{SDAE loss}
    \label{fig:sdae_loss}
\end{figure}

From right to left of Figure \ref{fig:sdae_loss}, the reconstruction, supervised, and combination of those two losses can be observed. 
Each loss function value decreases implying successful learning of the input distribution into a reduced dimensional latent space.
The reconstruction loss, alike the Simple and Deep AE shows implies a successful learning of a reduced dimensional latent space. 
The supervised loss decreasing implies that the model is able to properly credit labeling of anomalous data in order to refine the representation of the underlying distribution. 
When the encoded latent representation was printed out, higher values were observed in the first position, z: (z,\_).
This was expected as an emphasis on anomalous data was driven by the supervised loss with the label (1,0) for anomalies. 

Since this model was the focus, a separate learning cycle of 5000 epochs was performed.
This still did not result in any convergence to a solution and also did not offer any clue as to how many more epochs were necessary to find a solution. 

{}

\subsection{Prediction}

The test set involved two separate subsets of the original dataset: Building 408 and Building 733. The breakdown of anomalies shown in Table \ref{tab:total_anomalies}.

\begin{table}[h]
    \centering
    \begin{tabular}{|c|c|c|}
        \hline
        Subset Dataset & Total & Anomalies \\
        \hline
        Building 408 & 8782 & 273 \\
        \hline
        Building 733 & 8784 & 593 \\
        \hline
    \end{tabular}
    \caption{Anomalies in Test Datasets}
    \label{tab:total_anomalies}
\end{table}{}

The approach to finding anomalies involves procuring a loss distribution over each of the test datasets and implementing appropriate thresholds.
If a loss were to be below or exceed these thresholds then they were classified as anomalous. 
Determining the thresholds was done manually through trial and error. 

Table \ref{tab:loss_dist} shows the loss distribution for Building 408 and Building 733 respective to the model.
The Building 408 loss distribution required a lower threshold in order to capture anomalies - without it, they would go missed. 
This lower threshold did not contribute any meaningful effect for Building 733 except if it was set too high, it would increase the number of false positives without changing the number of true positives. 

\begin{table}[h]
    \centering
    \begin{tabular}{|c|c|c|}
        \hline
        Model & Building 408 & Building 733 \\
        \hline
        Simple AE &
        \subcaptionbox{Simple AE Loss Distribution for Blg 408 \label{1}}{\includegraphics[width = 2in]{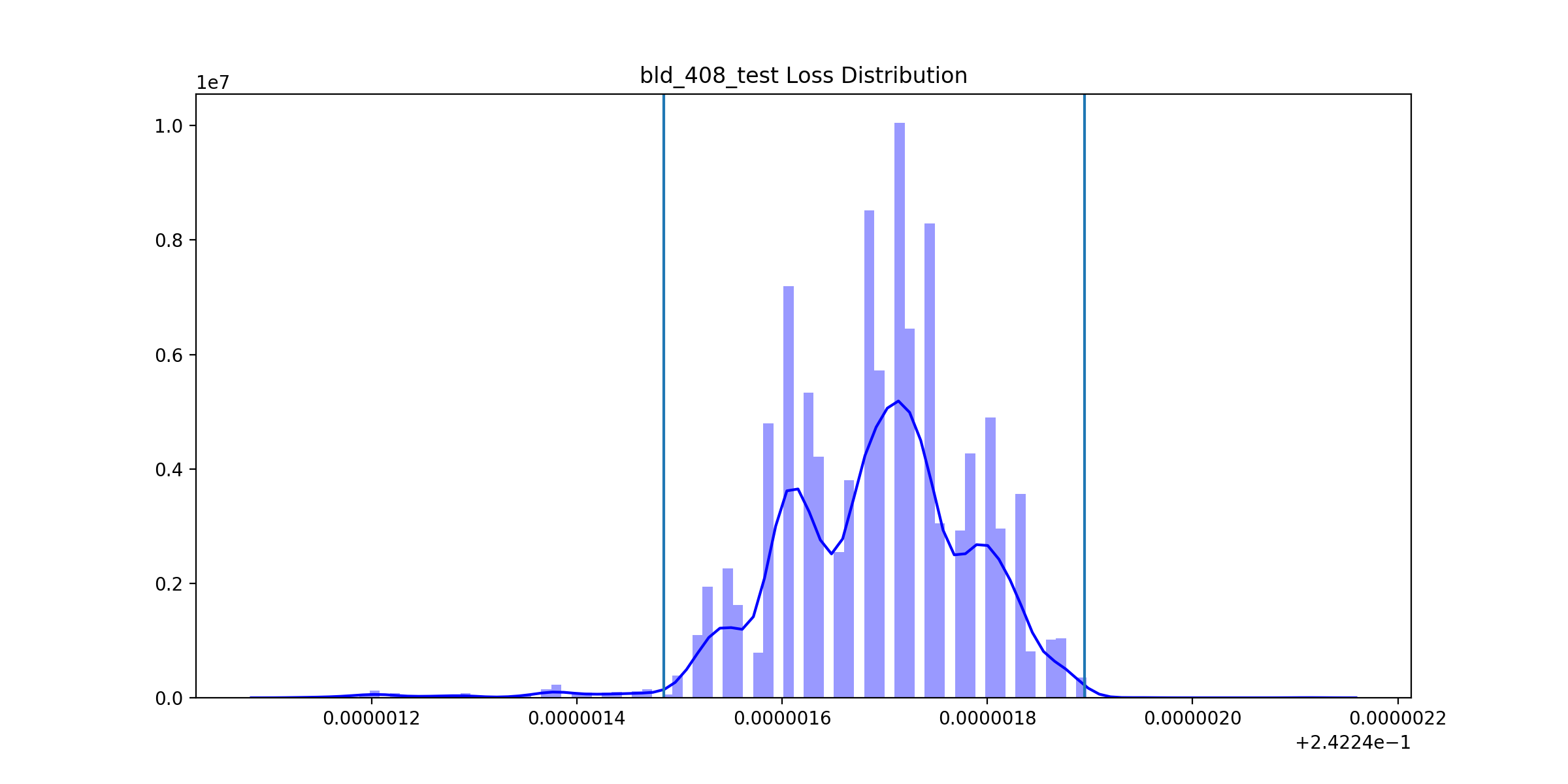}} &  
        \subcaptionbox{Simple AE Loss Distribution for Blg 733\label{1}}{\includegraphics[width = 2in]{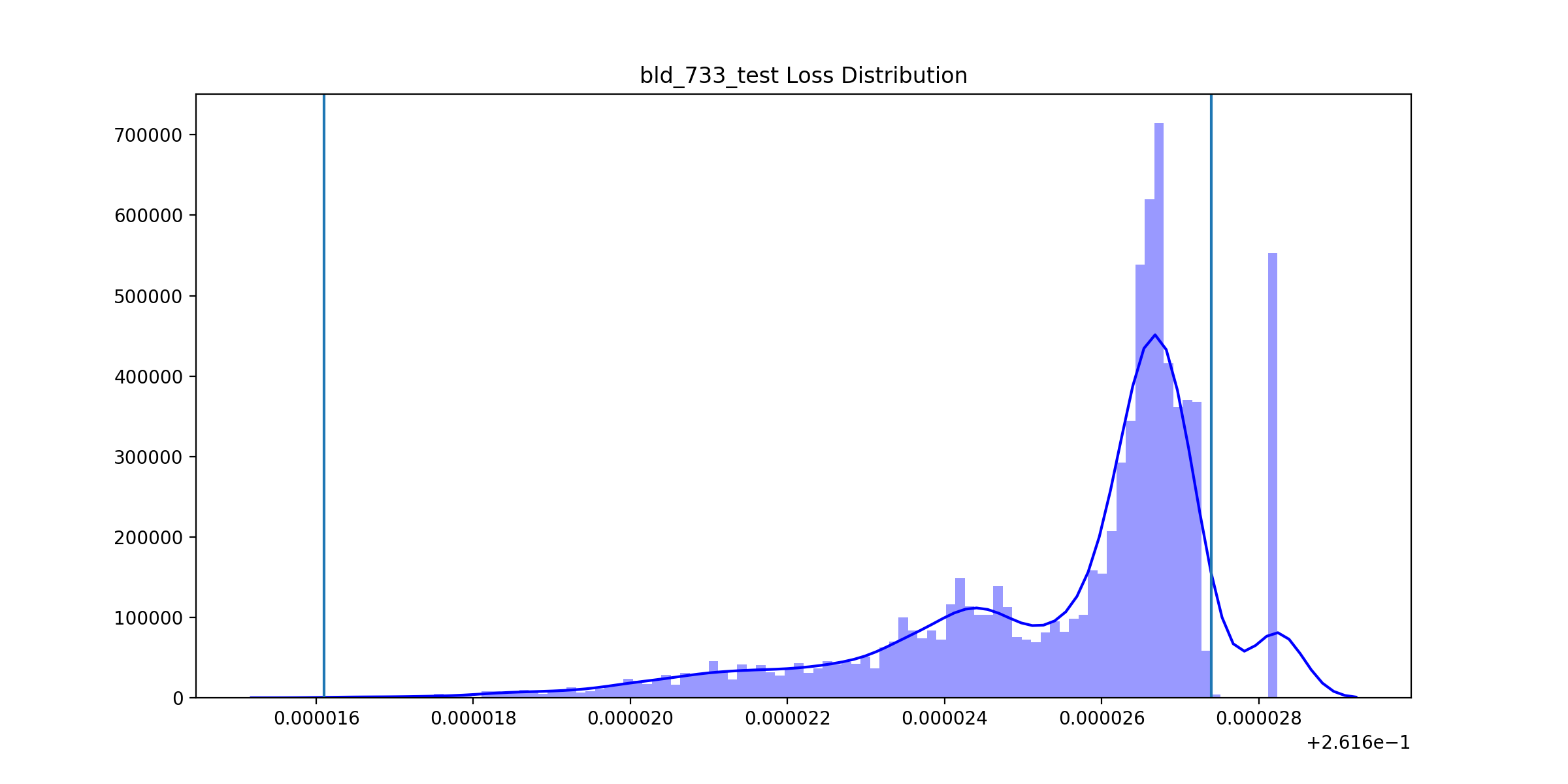}}\\
        \hline
        Deep AE &
        \subcaptionbox{Deep AE Loss Distribution for Blg 408 \label{1}}{\includegraphics[width = 2in]{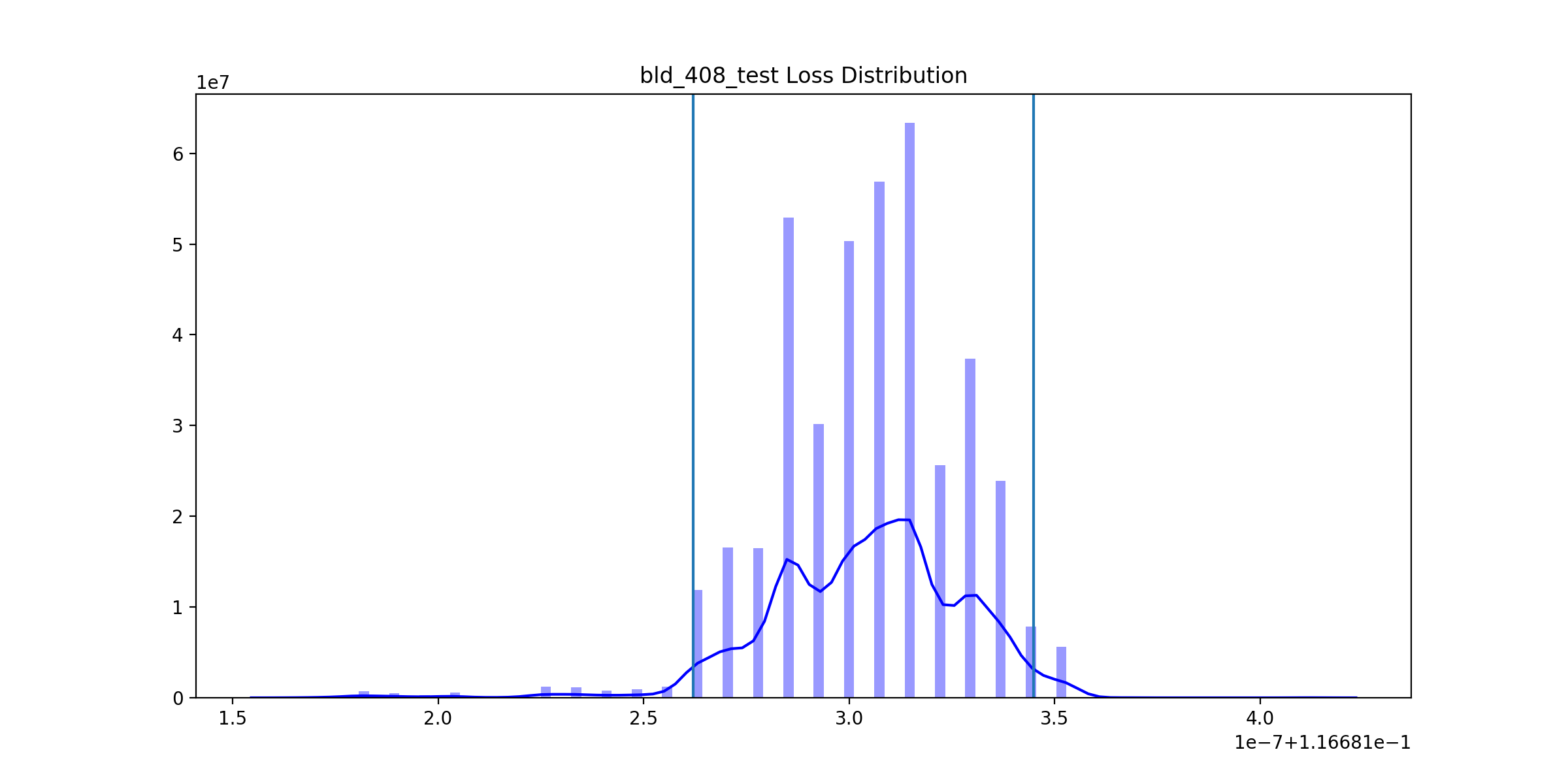}} &  
        \subcaptionbox{Deep AE Loss Distribution for Blg 733\label{1}}{\includegraphics[width = 2in]{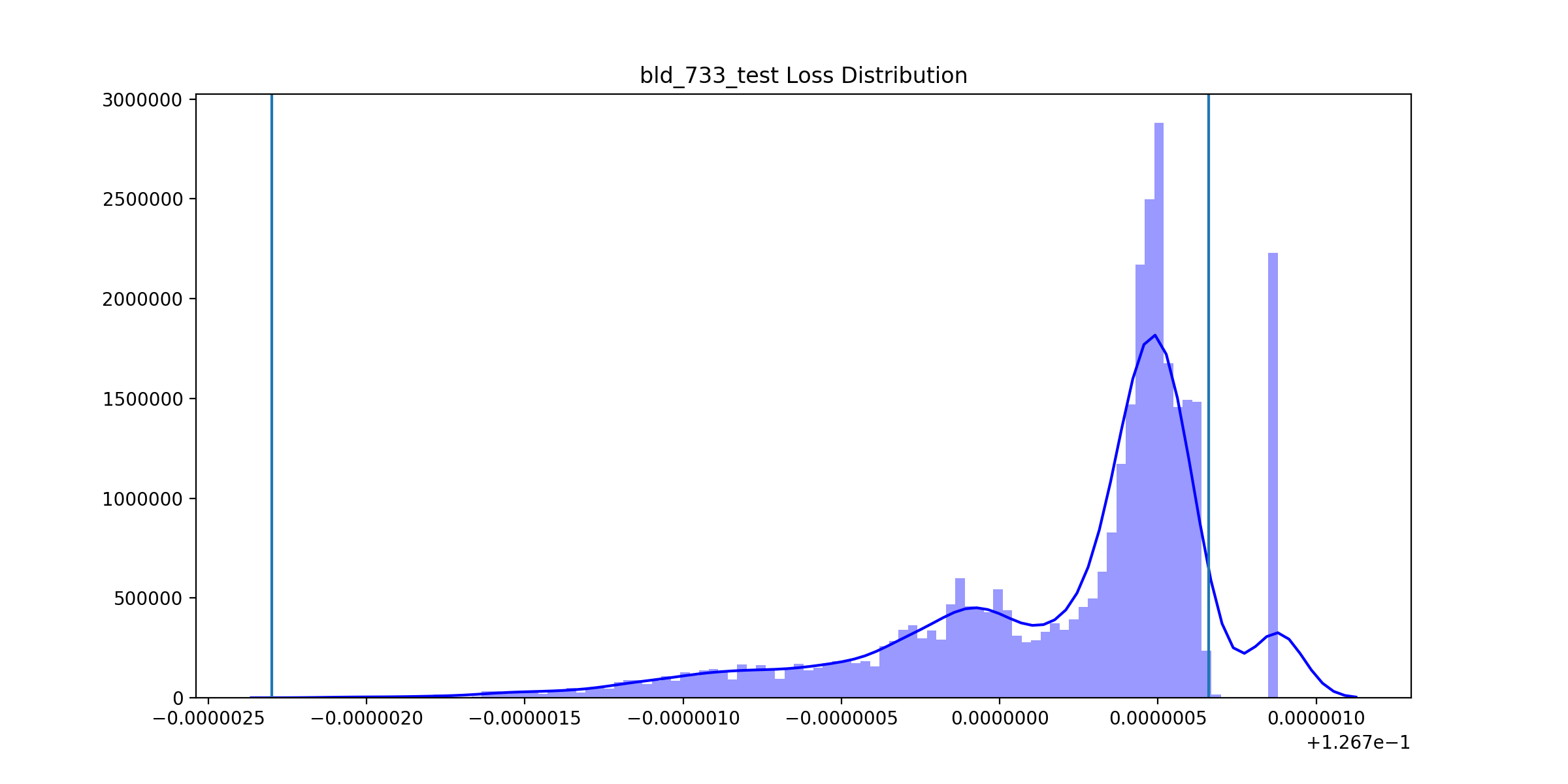}}\\
        \hline
        SDAE &
        \subcaptionbox{SDAE Loss Distribution for Blg 408 \label{1}}{\includegraphics[width = 2in]{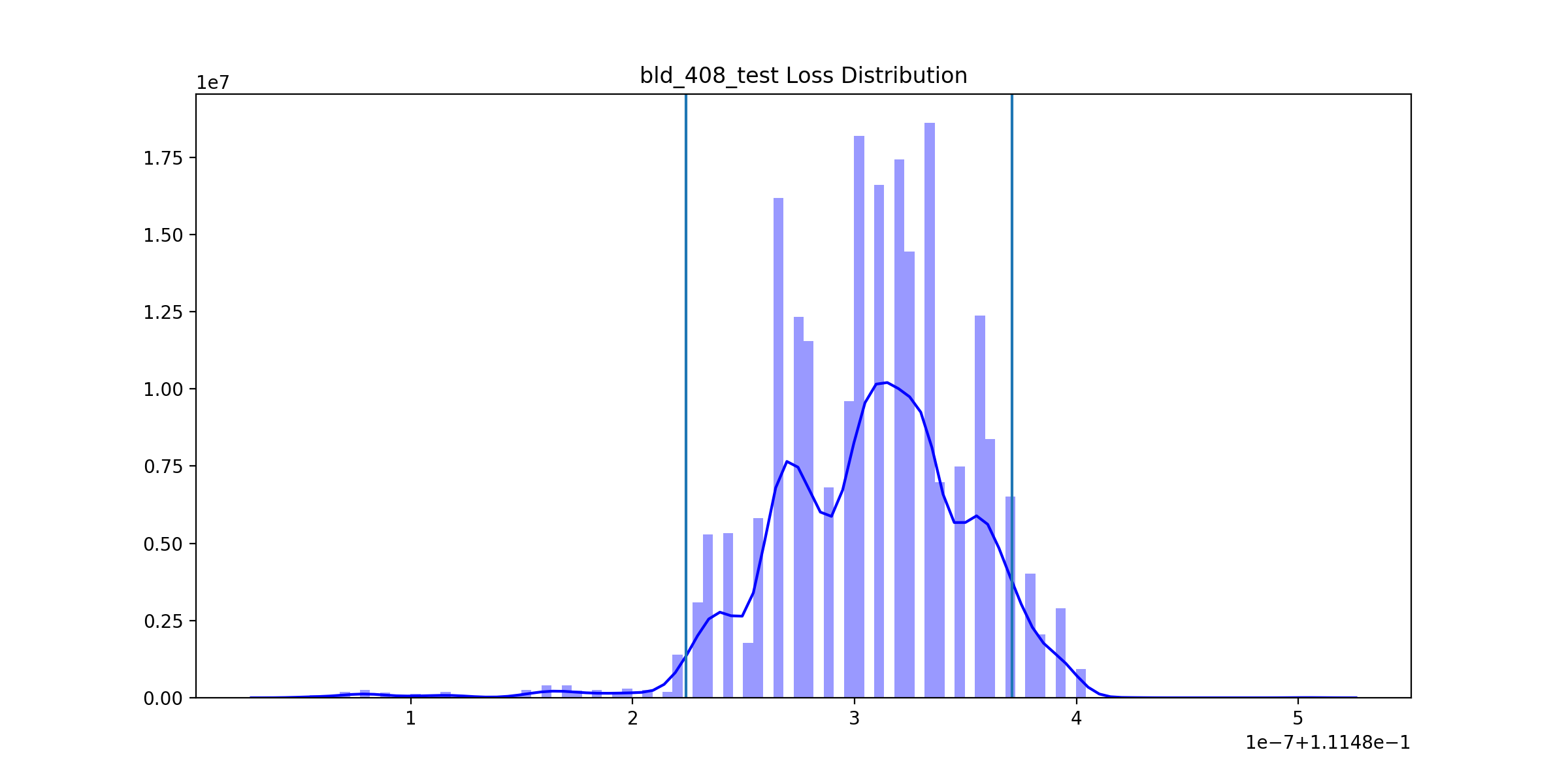}} &  
        \subcaptionbox{SDAE Loss Distribution for Blg 733\label{1}}{\includegraphics[width = 2in]{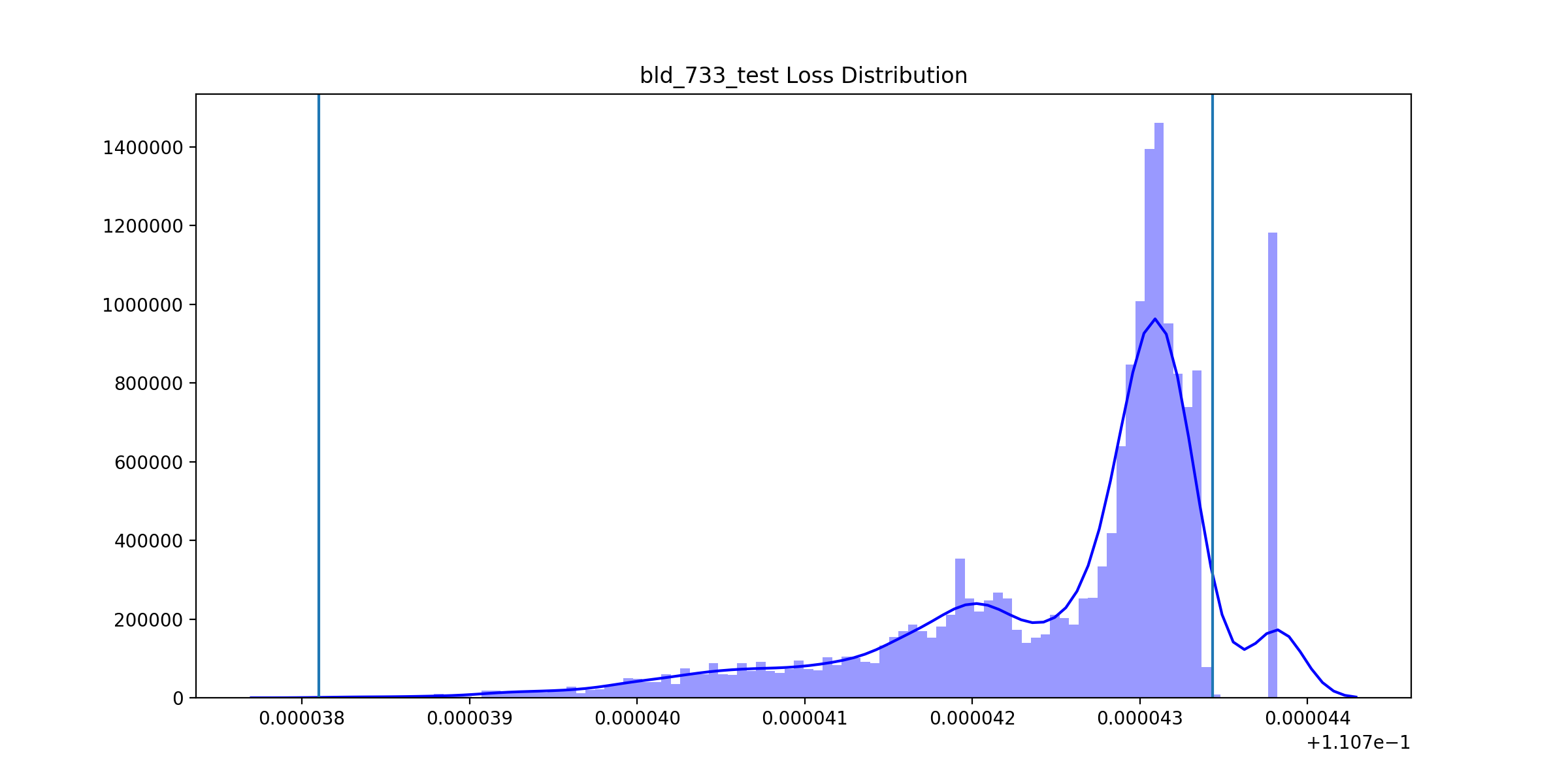}}\\
        \hline
    \end{tabular}
    \caption{Loss Distribution}
    \label{tab:loss_dist}
\end{table}{}

\subsection{Summary of Prediction Results}

Table \ref{tab:prediction_results_408} shows the results of each models prediction over the Building 408 labeled dataset. 
It is important to note that if the lower threshold for SDAE was increased by 1e-7 then the number of true positives increased to 162 (the most of the three), however the number of false positives also increased to 455. 

\begin{table}[h]
    \centering
    \begin{tabular}{|c||c||c||c|}
        \hline
        Criteria & Simple AE & Deep AE & SDAE\\
        \hline
        Upper Threshold & 0.242242 & 0.1166813449 & 0.11148037649 \\
        \hline
        Lower Threshold & 0.24224148 & 0.116681262 & 0.1114802211\\
        \hline
        Detected & 147 & 292 & 551 \\
        \hline
        True Positives & 147 & 157 & 153 \\
        \hline
        False Positives & 0 & 135 & 398 \\
        \hline
        Missed & 126 & 116 & 120\\
        \hline
    \end{tabular}
    \caption{Prediction Outcome Building 408}
    \label{tab:prediction_results_408}
\end{table}{}

Table \ref{tab:prediction_results_733} shows the results of each models prediction over the Building 733 labeled dataset. 

\begin{table}[ht!]
    \centering
    \begin{tabular}{|c||c||c||c|}
        \hline
        Criteria & Simple AE & Deep AE & SDAE\\
        \hline
        Upper Threshold & 0.2616274 & 0.12670066 & 0.1107434332 \\
        \hline
        Lower Threshold & 0.2616161 & 0.1266977 & 0.1107381 \\
        \hline
        Detected & 593 & 593 & 593 \\
        \hline
        True Positives & 593 & 593 & 593 \\
        \hline
        False Positives & 0 & 0 & 0 \\
        \hline
        Missed & 0 & 0 & 0 \\
        \hline
    \end{tabular}
    \caption{Prediction Outcome Building 733}
    \label{tab:prediction_results_733}
\end{table}{}

\section{Discussion}

The loss distribution for Building 733 showed a clear distinction between non-anomalous and anomalous datapoints as seen on the right side of each plot in Table \ref{loss_dist}. 
There existed an upper threshold that clearly identified 100\% of anomalies for that dataset with each model used. 
This implies that the underlying distribution of the metered values relative to the features were effectively learned by the model.

Building 408 dataset proved to be more difficult in classifying anomalies. 
The Simple AE captured 53\% of anomalies, Deep AE 57.5\%, and the SDAE captured 56\%. 
However, if the SDAE lower threshold increased (by 1e-7) then it could have detected 59\% of anomalies at the risk of also increasing the number of false positives captured. 
Although the SDAE performed marginally worse than the Deep AE in terms of anomalies detected, the number of false positives were accounted for. 
This means that the Deep AE outperformed both the Simple AE and SDAE after 1000 epochs of training. 
Even if we took the SDAE with it's higher anomaly detection rate (162), the number of false positives (455) would still put it at a lower performance compared to the Deep AE. 
The balance of these thresholds suggests that the losses associated with anomalies and non-anomalies were too close together to be properly classified. 
This could be because the Building 408 dataset might have been too far away from the training buildings dataset or possibly that the Building 408 itself held unique properties/anomalies unseen elsewhere in the source dataset.
If Building 408 was too far away from the training data, then it would be observed that the models would not be able to properly distinguish anomalies from non-anomalous data. 
Since this occurred with each variation of AE, it is highly likely that this is the case.
Assuming that this is the case, the number of false positives can be considered less meaningful and given that the SDAE has the highest detection of true positives given its higher lower threshold, then the SDAE outperforms the other two models with 162 true positives (59\% of total anomalies). 

It is worth reiterating that the goal of the project is to remove anomalies in order to improve the usefulness of the data. 
Further consideration should be made when using an autoencoder for anomaly detection on what the impact of false positives compared to missed anomalies is. 
If an anomaly imparts a much larger loss of information than the removal of a correct data point, then the autoencoder can be tuned to maximize the number of anomalies detected at the expense of false positives. 
This means that the SDAE outperformed both the Simple and Deep AE in recognizing anomalies. 

Both the training and test datasets were labeled manually by an expert, however, it must be considered that anomalies were missed. 
This could explain the high values of false positives for Building 408 and having such a highly accurate threshold to distinguish between anomalous and non-anomalous data. 
The models could, in theory, outperform that of a domain expert, and the false positive scores skewed.

The loss values observed with the SDAE at the prediction stage are lower than both the Simple and Deep AE. 
Despite being trained over an equal time frame, the loss of the SDAE implies that it was able to achieve a more accurate representation of the the underlying input distribution. 
It could be argued that given enough time, a more distinct solution could be achieved by the SDAE. 

\section{Conclusion}

The issues surrounding Building 408 with it's low succession of detecting anomalies gears the problem toward the chosen subset and source dataset. 
Questioning the dataset itself is why this problem exists - the quality of the data used is always the biggest indicator of the success of a project. 
With the results and thresholds given, the Deep AE performs the best - however taken into light with the possible imperfections of the Building 408 dataset, as well as the meaningfulness of false positives to missed anomalies, the SDAE performs the best in detecting anomalies. 
For an initial investigation into anomaly detection using Autoencoders, these results are promising as a viable method of application use cases. 

\section{Future Work}

Since the thresholds were done manually through trial and error, work into calculating these thresholds dynamically that maximized true positives, while minimizing false positives and missed true positives would make this application much more effective.  
The SDAE lower threshold could be further investigated into an exact split that could potentially separate the true positives from the false positives. 

Further work into the subsets used to train and test the model is necessary to determine if the quality of subsets is representative enough of the entire distribution. 
That being said, the more important piece of future work in this realm would be to label the entire source dataset for training, validation, and testing purposes. 

If the models were each trained for a longer period of time until each could converge to a solution and then compared in performance, more detailed or intuitive results could be evaluated. 

\bibliographystyle{unsrt}
\bibliography{anomalydetection.bib}

\end{document}